\documentclass{llncs}
\pdfoutput=1

\usepackage{amsmath}
\usepackage{microtype}

\usepackage[round,sort,sectionbib]{natbib}

\usepackage{listings}
\lstdefinelanguage{Grasp} {
        morekeywords={architecture, template, quality_attribute, requirement, ra
tionale, reason, supports, inhibits, extends},
        morekeywords={system, layer, over, component, connector, requires, provi
des, link, to, check, property},
        morekeywords={over, realises, subsetof, accepts, true, false},
        sensitive=true,
        morestring=[b]",
        morestring=[b]',
        morecomment=[l]{//},
        morecomment=[s]{/*}{*/},
        captionpos=b,
        tabsize=2,
        basicstyle=\sffamily\scriptsize
}

\lstdefinelanguage{Minion} {
        morekeywords={MINION 3, BOOL, DISCRETE, VARIABLES, CONSTRAINTS,
        SEARCH, EOF, reify, watched, or, eq, reifyimply, and}
        sensitive=true,
        captionpos=b,
        tabsize=2,
        basicstyle=\sffamily\scriptsize
}

\title{Modelling Constraint Solver Architecture Design as a Constraint Problem}

\author{Ian Gent \and Chris Jefferson \and Lars Kotthof\/f \and Ian Miguel\\
\email{\{ipg,caj,larsko,ianm\}@cs.st-andrews.ac.uk}}
\institute{University of St Andrews}

\begin{document}

\maketitle

\begin{abstract}
Designing component-based constraint solvers is a complex problem. Some
components are required, some are optional and there are interdependencies
between the components. Because of this, previous approaches to solver design
and modification have been ad-hoc and limited.  We present a system that
transforms a description of the components and the characteristics of the target
constraint solver into a constraint problem. Solving this problem yields the
description of a valid solver.  Our approach represents a significant step
towards the automated design and synthesis of constraint solvers that are
specialised for individual constraint problem classes or instances.
\end{abstract}

\section{Introduction}

Most current constraint solvers, such as Minion \citep{minion}, are constructed
to be as general as possible. They are monolithic in design, accepting a broad
range of models. While this generality is convenient, it leads to a complex
internal architecture, resulting in significant overheads and inhibiting
efficiency, scalability and extensibility. Another drawback is that current
solvers perform little analysis of an input model, so the features of an
individual model cannot be exploited to produce a more efficient solving
process. To mitigate these drawbacks, constraint solvers often allow manual
tuning of the solving process. However, this requires considerable expertise,
preventing the widespread adoption of constraint solving.

A possible solution to this problem is to automatically generate specialised
constraint solvers. A given problem class or instance is analysed and the most
suitable solver components identified. These components are then assembled into
a solver. The artificial intelligence community has shown a lot of interest in
this problem recently, especially in the context of algorithm
portfolios \citep{satzilla,cphydra,gent10lazy,algportfolios}. The solution
proposed above takes the portfolio idea a step further -- instead of selecting
or configuring an existing solver, we aim to \emph{synthesise} a specialised
solver.

The techniques for analysing problems and identifying the most suitable solution
strategies are still applicable for our approach, but in addition we are faced
with the difficult problem of automatically assembling a constraint solver from
a list of components and a specification of the problem it is to solve.
Constraint programming itself is a natural fit for solving this configuration
problem. In the remainder of this paper, we detail a way of expressing the
architecture of a constraint solver in such a way that it can be solved as a
constraint problem.

\section{Background}

Several other approaches for generating specialised constraint solvers exist.
The \textsc{Multi-tac} \citep{multitac} system configures and compiles a
constraint solver for a specific set of problems. It does not synthesis a new
constraint solver from a library of components, but customises a base solver.
The customisations are limited to heuristics and do not affect all parts of the
constraint solver.

\textsc{KIDS} \citep{kids} is a more general system and
synthesises efficient algorithms from an initial specification. The approach is
knowledge-based, i.e.\ the user supplies the knowledge required to generate an
efficient algorithm for the specific problem. Refinements are limited to a
number of generic transformation operations and again are not capable of
customising all parts of a solver.

A review of the literature on analysing combinatorial problems and selecting the
most efficient way of solving them is beyond the scope of this paper. Some of
the most prominent systems are SATzilla \citep{satzilla} and
CPHydra \citep{cphydra}, which select from a portfolio of SAT and constraint
solvers respectively based on the characteristics of a problem using machine
learning.

\medskip

The synthesis of a constraint solver from components is a configuration problem,
many instances of which have been discussed in the literature. An overview can
be found in the Configuration chapter of the Handbook of Constraint
Programming \citep{handbook}.

One of the earliest approaches to solving configuration problems as constraint
problems is by \cite{mittal_dynamic_1990} and proposes dynamic constraint
problems that introduce new variables as the requirements for configured
components become known. They furthermore require special constraints that
express whether a variable is still relevant to the partially solved problem
based on the assignments made so far.

\cite{sabin_configuration_1996} propose solving configuration
problems as composite constraint satisfaction problems where values for
variables can be constraint problems themselves.
\cite{stumptner_generative_1998} introduce the constraint-based configuration
system COCOS. Their system requires several extensions of the standard
constraint paradigm as well. \cite{mailharro_classification_1998}
proposes a constraint formulation that integrates concepts from object-oriented
programming. His approach relies on many of the concepts introduced in earlier
work and infinite-sized domains for variables. \cite{hinrich_using_2004} use
object-oriented constraint satisfaction for modelling configuration problems.
They then transform the constraint model into first order logic sentences and
find a solution using a theorem solver.

Our approach works without the need to modify an existing off-the-shelf
constraint solver and  a solution gives a complete configuration of a solver.
There is no need to solve a series of refined constraint problems. This is
crucial for us because we are aiming to do this in the context of generating a
constraint solver that is specialised to solve a particular problem more
efficiently and want to keep possible overheads, such as repeatedly generating
constraint models and calling a constraint solver, as minimal as possible.

\section{Architecture specification}

We use the generic software architecture description language
GRASP \citep{wicsa,grasp} to describe the components of a constraint solver. The
advantages of using a generic architecture description language include
available tools for checking architecture descriptions for consistency and that
people without a background in constraint programming are able to work with it.
We chose GRASP because it is being developed by a research group at our
department and we are able to influence the design of the language towards
meeting the requirements for modelling constraint solvers.

A full description of GRASP is beyond the scope of this paper and not necessary
for our purposes. The relevant elements of the language are described below.
\begin{description}
\item[templates] Templates are the high-level elements of the language that
    describe components. A single template can describe a memory manager for
    example. Templates may take parameters when they are instantiated to
    customise their behaviour further.
\item[requires/provides] Describe things a template needs and offers for other
    templates to use. A memory manager for example provides a facility for
    storing and retrieving data. This facility could be required by a variable
    to keep track of its domain.
\item[properties] Properties characterise components beyond the generic
    facilities they provide. A Boolean variable for example would have the
    property that the size of the domain is at most two.
\item[checks] Check statements model the interdependencies between components
    and restrictions of customisations of a component. A component that
    implements a specific constraint for example would place restrictions on the
    parameters it can be customised with (i.e.\ the variables that it
    constrains) by e.g.\ limiting the domain size.
\end{description}

The check statements of GRASP provide much power and flexibility. Only a small
subset of this is needed to express the components of a constraint solver
though. The relevant parts are explained below.
\begin{description}
\item[A subsetof B] Asserts that set B contains all the elements of set A. It is
    used to ensure that a certain implementation has a specific set of
    properties and provides a specific set of facilities.  It can also be used
    to ensure that an implementation does \emph{not} have a property or
    facility.
\item[A accepts B] Asserts that B is accepted as A, e.g.\ if A
    is the parameter given to the implementation of a constraint and B is a
    variable implementation, it makes sure that the constraint can be put on
    variables of that type.
\end{description}

Apart from the components that describe the building blocks of a solver, there
is a top-level meta-component that describes the problem to be solved. It
specifies the types of variables and constraints needed and which constraint
implementation needs to work with which variable implementation.

The description of the constraint solver consists of a library of solver
components specified this way and the problem meta-component. The library of
solver components is not specific to any constraint problem to be solved by the
generated solver and describes all the implementation options for any solver.
The problem meta-component encodes the requirements for solving a particular
constraint problem and links components from the library into an actual
constraint solver.

\section{Constraint model}

The requirements of a component naturally map to variables in a constraint
problem that we want to find assignments for. The domain of each of those
variables is determined by the components which provide the facility required,
i.e.\ the possible implementations. Each implementation variable has a set of
provides and properties attached to it. The set of provides is necessary because
an implementation may provide more than the one main facility that would be
required by another component.  If a variable is assigned a value which
determines its implementation, it \emph{must} provide all the facilities and
have all the properties that this implementation provides and has and it
\emph{must not} provide any other facilities or have any other properties. We
therefore add constraints to ensure that a component variable has a certain
property or provide if and only if it is assigned an implementation that has
this property or provide.

There are several cases we need to consider for converting the check statements
of GRASP into constraints. The first case is of the form \texttt{list subsetof
properties/provides}. This requires a component implementation to provide a
list of facilities or have a set of properties. The translation into constraints
is straightforward; we simply require the things in \texttt{list} to be in the
set of \texttt{properties/provides}. The second case of the form
\texttt{properties/provides subsetof list}. This is the opposite of the previous
case and \emph{forbids} the properties/provides which are not listed explicitly.
The translation into constraints is analogous to the previous case.

The final case deals with the \texttt{accepts}. The general requirement encoded
is that if a parameter to an implementation requires a certain property or
facility, the implementation of the parameter must provide it. The corresponding
constraints are implications that require properties and provides of an
implementation that might be used as a parameter to be set if they are set for
the parameter.

\subsection{Conditional variables and constraints}

The variables and constraints mentioned so far are only valid at the top level,
i.e.\ for the problem meta-component.  We need additional constructs that encode
the requirements that arise if a component is implemented in a certain way. The
variables and constraints to encode the requirements take the same form as
above, but they have prerequisites that need to be true in order for them to
become relevant.

We chose an explicit representation of the prerequisites where the conditional
variables encode them in their names. The names of the variables that model the
requirements for an implementation of a component not at the top level are
prefixed by the implementation choices for the top-level components. The
constraints on these variables can be encoded as an implication, e.g.\ if
component x is implemented as an A, its first parameter needs to have property
Y. The name of the variable that models this first parameter would have a prefix
that indicates that the superior component x is implemented as an A. The
left-hand side of the implication is a conjunction of the implementation
decisions made in the prerequisites.

\subsection{Modelling language}\label{minionmod}

We decided to use the modelling language of the Minion constraint solver. While
it would be easier to use a more high-level language such as Essence, we need
more fine-grained control over the solving process. In particular, we need to be
able to specify the order of variables and values in the domains of variables to
guide the solver towards the implementations we consider the most suitable
ones. This enables us to analyse the constraint problem to be solved with the
synthesised solver, identify the component implementations that are likely to
provide the best performance and encode this in the constraint problem through
the variable and value orderings.

The decision to use the Minion input language has some ramifications for the
model. First, Minion only supports integer domain values and all component
implementations, properties and provides must be mapped to integers.
Furthermore, some of the constraints that the model uses are not provided by
Minion and must be encoded with additional constraints and variables.

For the provides and the properties of each component variable, we added an
auxiliary array of Boolean variables to represent the set. If the $i$th Boolean
is set to true, the $i$th property or provides is present in the set. This means
that two auxiliary arrays of Booleans are added for each component variable.

Almost all constraints can be encoded directly in Minion. We used the
\texttt{watched-or} constraint to express that a component variable can
have a property or provides if and only if it is assigned one of the
implementations that have it. The conjunction to encode the conditional
constraints was implemented with a \texttt{watched-and}. The only constraints
which cannot directly be translated into Minion are the implications, as Minion
only allows implications between a Boolean variable and a constraint. To
mitigate this, we introduced channelling variables, one for each auxiliary array
of Booleans that encode the properties and provides. The left hand side of the
implication is linked to the channelling variable through an if and only if
(\texttt{reify} in Minion) and the right hand side is connected to the
channelling variable by an implication constraint (\texttt{reifyimply} in
Minion).

\section{Example}

Consider the constraint problem below.
\begin{align*}
    pvx + pvy &= pvw + pvc6\\
    pvx &= pvz
\end{align*}

The GRASP specification of a solver component library and problem meta-component
that corresponds to the constraint problem are shown in Figure~\ref{grasp}.  The
problem meta-component requires five variables and two constraints with certain
properties and restrictions. The variable and constraint implementations impose
further restrictions and may in turn require a memory manager. Constant
variables have domain size 1, Boolean variables domain size 2 and discrete
variables arbitrary-sized domains. A GAC sum needs to be able to remove values
from the domain of the variables it constrains while a Boolean sum needs its
first argument to be a Boolean (domain size 2) and its second argument to be a
constant variable (domain size 1). The memory manager does not have any special
properties or further requirements.

\begin{figure}
\begin{lstlisting}[firstline=2,language=Grasp]
architecture Solver {
  template CopyMemoryFactory() {
    provides IMemoryManager;
    provides IRawMemory;
    provides IViewHistory;
  }
  template DiscreteVariableFactory() {
    provides IPropVariable;
    provides IRemoveFromDomain;
    requires IMemoryManager domain;
    requires IMemoryManager bounds;
    property domainType = "bound";
    check domain.getProperties() subsetof [(MemoryChanges, 'Single')];
    check bounds.getProperties() subsetof [];
  }
  template BoolVariableFactory() {
    provides IPropVariable;
    provides IRemoveFromDomain;
    requires IMemoryManager bounds;
    property domainType = "bound";
    check bounds.getProperties() subsetof [(MemoryChanges, 'Single')];
    property domainSize = 2;
  }
  template ConstantVariableFactory() {
    provides IPropVariable;
    provides IRemoveFromDomain;
    property domainSize = 1;
    property domainType = "bound";
  }
  template GACSumFactory(P1, P2) {
    provides ISumEqCon;
    requires IMemoryManager m;
    check m.getProperties() subsetof [];
    check [IRawMemory] subsetof m.getInterfaces();
    check [IRemoveFromDomain,IPropVariable] subsetof P1.getInterfaces();
    check [IRemoveFromDomain,IPropVariable] subsetof P2.getInterfaces();
  }
  template BoolSumFactory(P1, P2) {
    provides ISumEqCon;
    requires IMemoryManager m;
    check m.getProperties() subsetof [];
    check [(domainSize, 2)] subsetof P1.getProperties();
    check [(domainSize, 1)] subsetof P2.getProperties();
    check [IPropVariable] subsetof P1.getInterfaces();
    check [IPropVariable] subsetof P2.getInterfaces();
  }
  template ThisProblem() {
    provides IProblem;
    requires IPropVariable pvw, pvx, pvy, pvz, pvc6;
    requires ISumEqCon scA, scB;
    check pvx.getProperties()  subsetof
        [(domainType, 'bound'), (domainSize, 2)];
    check pvy.getProperties()  subsetof
        [(domainType, 'bound')];
    check pvc6.getProperties() subsetof
        [(domainType, 'bound'), (domainSize, 1)];
    check scA.param(1) accepts (pvx  +
        [(domainType, 'bound'), (domainSize, 2)]);
    check scA.param(1) accepts (pvy  +
        [(domainType, 'bound')]);
    check scA.param(2) accepts (pvw  +
        [(domainType, 'bound'), (domainSize, 1)]);
    check scA.param(2) accepts (pvc6 +
        [(domainType, 'bound'), (domainSize, 1)]);
    check scB.param(1) accepts (pvx  +
        [(domainType, 'bound'), (domainSize, 2)]);
    check scB.param(1) accepts pvz;
  }
}
\end{lstlisting}
    \caption{Solver architecture description for simple constraint problem.}
    \label{grasp}
\end{figure}

\begin{figure}
\begin{lstlisting}[firstline=2,language=Minion]
MINION 3
**VARIABLES**
DISCRETE pvw_1_domain {1..1}
BOOL pvw_1_domain_properties[3]
BOOL pvw_1_domain_provides[7]
DISCRETE pvw_1_bounds {1..1}
BOOL pvw_1_bounds_properties[3]
BOOL pvw_1_bounds_provides[7]
DISCRETE pvw_2_bounds {1..1}
BOOL pvw_2_bounds_properties[3]
BOOL pvw_2_bounds_provides[7]
DISCRETE pvw {1..3}
BOOL pvw_properties[3]
BOOL pvw_provides[7]
...
**CONSTRAINTS**
reify(watched-or({eq(pvw_1_domain, 1)}), pvw_1_domain_provides[0])
eq(pvw_1_domain_provides[6], 0)
eq(pvw_1_domain_provides[3], 0)
reify(watched-or({eq(pvw_1_domain, 1)}), pvw_1_domain_provides[1])
eq(pvw_1_domain_provides[4], 0)
eq(pvw_1_domain_provides[5], 0)
reify(watched-or({eq(pvw_1_domain, 1)}), pvw_1_domain_provides[2])
eq(pvw_1_domain_properties[2], 0)
eq(pvw_1_domain_properties[1], 0)
eq(pvw_1_domain_properties[0], 0)
reify(watched-or({eq(pvw_1_bounds, 1)}), pvw_1_bounds_provides[0])
eq(pvw_1_bounds_provides[6], 0)
eq(pvw_1_bounds_provides[3], 0)
reify(watched-or({eq(pvw_1_bounds, 1)}), pvw_1_bounds_provides[1])
eq(pvw_1_bounds_provides[4], 0)
eq(pvw_1_bounds_provides[5], 0)
reify(watched-or({eq(pvw_1_bounds, 1)}), pvw_1_bounds_provides[2])
eq(pvw_1_bounds_properties[2], 0)
eq(pvw_1_bounds_properties[1], 0)
eq(pvw_1_bounds_properties[0], 0)
reify(watched-or({eq(pvw_2_bounds, 1)}), pvw_2_bounds_provides[0])
eq(pvw_2_bounds_provides[6], 0)
eq(pvw_2_bounds_provides[3], 0)
reify(watched-or({eq(pvw_2_bounds, 1)}), pvw_2_bounds_provides[1])
eq(pvw_2_bounds_provides[4], 0)
eq(pvw_2_bounds_provides[5], 0)
reify(watched-or({eq(pvw_2_bounds, 1)}), pvw_2_bounds_provides[2])
eq(pvw_2_bounds_properties[2], 0)
eq(pvw_2_bounds_properties[1], 0)
eq(pvw_2_bounds_properties[0], 0)
eq(pvw_provides[0], 0)
eq(pvw_provides[6], 0)
reify(watched-or({eq(pvw, 1), eq(pvw, 2), eq(pvw, 3)}), pvw_provides[3])
eq(pvw_provides[1], 0)
reify(watched-or({eq(pvw, 1), eq(pvw, 2), eq(pvw, 3)}), pvw_provides[4])
eq(pvw_provides[5], 0)
eq(pvw_provides[2], 0)
reify(watched-or({eq(pvw, 3)}), pvw_properties[2])
reify(watched-or({eq(pvw, 2)}), pvw_properties[1])
reify(watched-or({eq(pvw, 1), eq(pvw, 2), eq(pvw, 3)}),
    pvw_properties[0])
...
reify(watched-and({eq(scA, 1)}), scA_1_param_1_provides_channel[3])
reifyimply(eq(scA_param_1_provides[3], 1),
    scA_1_param_1_provides_channel[3])
reify(watched-and({eq(scA, 1)}), scA_1_param_1_provides_channel[4])
reifyimply(eq(scA_param_1_provides[4], 1),
    scA_1_param_1_provides_channel[4])
reify(watched-and({eq(scA, 1)}), scA_1_param_2_provides_channel[3])
reifyimply(eq(scA_param_2_provides[3], 1),
    scA_1_param_2_provides_channel[3])
reify(watched-and({eq(scA, 1)}), scA_1_param_2_provides_channel[4])
reifyimply(eq(scA_param_2_provides[4], 1),
    scA_1_param_2_provides_channel[4])
...
**SEARCH**
**EOF**
\end{lstlisting}
    \caption{Excerpts of constraint model for Figure~\ref{grasp}.}
    \label{minion}
\end{figure}

Parts of the Minion model generated from this description is shown in
Figure~\ref{minion}.  The first part shows the variables generated for the
requirement \texttt{IPropVariable pvw} in the GRASP model (Figure~\ref{grasp}).
Apart from the main variable, there are auxiliary variables for the properties
and the provides as well as variables which model the conditional requirements
of \texttt{pvw} being implemented in a particular way.

The first section of the constraints section models the properties and requires
\texttt{pvw} (or one of its requirements) will have if being implemented in a
particular way. We especially refer the reader to the last couple of lines
before the \ldots{} -- these express what possible implementations for
\texttt{pvw} would give it specific properties/provides and that some of the
provides are not given by any of the candidate implementations and are therefore
always not in the set (Boolean array element set to 0).

The second part of the constraints section models the check statements that
affect the parameters of the sum constraint implementations. The variables for
these components are not shown for space reasons, but are analogous to the
variables for the \texttt{pvw} component. A pair of Minion constraints is
required to model the implication of a component being implemented in a specific
way, as outlined in Section~\ref{minionmod}. The first constraint reifies the
conditions with the channelling variable while the second constraint establishes
the implication between the channelling variable and the actual property or
provide.

Note that some of the \texttt{check\ldots{}accepts} statements are given
additional properties for the variables to check in the GRASP model. Only the
properties which are not explicitly given need to be checked by the generated
constraints.

Minion finds a valid constraint solver architecture for the, admittedly trivial,
encoded problem (the first solution to the generated constraint model) in just a
couple of milliseconds.

\section{Limitations and future work}

A limitation of the current system is that we are unable to express
requirements which have a global effect on all components, such as whether to
attach debug information. At present, we are unable to express this in GRASP and
therefore cannot add it to the constraint model. We are planning on extending
GRASP to support this.

We have found that in practice while solving the generated constraint problems
for the first solution is quick, enumerating all solutions takes a long time
because of the auxiliary variables which result in sets of separate solutions
that specify the same constraint solver being found.

\section{Conclusions}

We have presented a way of encoding the configuration of the architecture of a
constraint solver as a constraint problem such that a solution to the problem
specifies a valid solver. This represents a major step towards automated
synthesis of constraint solvers from a library of components for a given
problem. Given a library and components and a problem specification, we can
automatically and efficiently synthesis a constraint solver.

Modelling the architecture of a constraint solver as a standard constraint
problem enables us to use off-the-shelf software to solve this complex
configuration problem using tried and tested techniques. Instead of a single
solver, we can easily generate all valid solvers by finding all solutions to the
configuration problem instead of only the first one.

\subsection*{Acknowledgements}

GRASP was designed by Dharini Balasubramaniam and Lakshitha de Silva. Peter
Nightingale was involved with the modifications of GRASP.  Lars Kotthof\/f is
supported by a SICSA studentship. This work was supported by EPSRC grant
EP/H004092/1. We thank the anonymous reviewer for their comments.

\bibliographystyle{plainnat}
\renewcommand\bibname{References}
\bibliography{solvercsp}

\begin{thebibliography}{15}
\providecommand{\natexlab}[1]{#1}
\providecommand{\url}[1]{\texttt{#1}}
\expandafter\ifx\csname urlstyle\endcsname\relax
  \providecommand{\doi}[1]{doi: #1}\else
  \providecommand{\doi}{doi: \begingroup \urlstyle{rm}\Url}\fi

\bibitem[Balasubramaniam et~al.(2011)Balasubramaniam, de~Silva, Jefferson,
  Kotthoff, Miguel, and Nightingale]{wicsa}
Dharini Balasubramaniam, Lakshitha de~Silva, Chris Jefferson, Lars Kotthoff,
  Ian Miguel, and Peter Nightingale.
\newblock Dominion: An architecture-driven approach to generating efficient
  constraint solvers.
\newblock In \emph{9th Working IEEE/IFIP Conference on Software Architecture
  (WICSA)}, 2011.

\bibitem[de~Silva and Balasubramaniam(2011)]{grasp}
Lakshitha de~Silva and Dharini Balasubramaniam.
\newblock A model for specifying rationale using an architecture description
  language.
\newblock Technical report, University of St Andrews, 2011.

\bibitem[Gent et~al.(2010)Gent, Jef\/ferson, Kotthof\/f, Miguel, Moore,
  Nightingale, and Petrie]{gent10lazy}
I.~Gent, C.~Jef\/ferson, L.~Kotthof\/f, I.~Miguel, N.~Moore, P.~Nightingale,
  and K.~Petrie.
\newblock Learning when to use lazy learning in constraint solving.
\newblock In \emph{ECAI}, pages 873--878, August 2010.

\bibitem[Gent et~al.(2006)Gent, Jefferson, and Miguel]{minion}
Ian Gent, Christopher Jefferson, and Ian Miguel.
\newblock {MINION}: A fast scalable constraint solver.
\newblock In \emph{Proceedings of the Seventeenth European Conference on
  Artificial Intelligence}, pages 98--102, 2006.

\bibitem[Gomes and Selman(2001)]{algportfolios}
Carla~P. Gomes and Bart Selman.
\newblock Algorithm portfolios.
\newblock \emph{Artif. Intell.}, 126\penalty0 (1-2):\penalty0 43--62, 2001.

\bibitem[Hinrich et~al.(2004)Hinrich, Love, Petrie, Ramshaw, Sahai, and
  Singhal]{hinrich_using_2004}
Tim Hinrich, Nathaniel Love, Charles Petrie, Lyle Ramshaw, Akhil Sahai, and
  Sharad Singhal.
\newblock Using {Object-Oriented} constraint satisfaction for automated
  configuration generation.
\newblock In \emph{{DSOM}}, 2004.

\bibitem[Mailharro(1998)]{mailharro_classification_1998}
Daniel Mailharro.
\newblock A classification and constraint-based framework for configuration.
\newblock \emph{Artif. Intell. Eng. Des. Anal. Manuf.}, 12:\penalty0 383--397,
  1998.

\bibitem[Minton(1996)]{multitac}
Steven Minton.
\newblock Automatically configuring constraint satisfaction programs: A case
  study.
\newblock \emph{Constraints}, 1:\penalty0 7--43, 1996.

\bibitem[Mittal and Falkenhainer(1990)]{mittal_dynamic_1990}
Sanjay Mittal and Brian Falkenhainer.
\newblock Dynamic constraint satisfaction problems.
\newblock In \emph{{AAAI}}, pages 25--32, 1990.

\bibitem[{O'Mahony} et~al.(2008){O'Mahony}, Hebrard, Holland, Nugent, and
  {O'Sullivan}]{cphydra}
E.~{O'Mahony}, E.~Hebrard, A.~Holland, C.~Nugent, and B.~{O'Sullivan}.
\newblock Using case-based reasoning in an algorithm portfolio for constraint
  solving.
\newblock In \emph{Irish Conference on Artificial Intelligence and Cognitive
  Science}, 2008.

\bibitem[Rossi et~al.(2006)Rossi, van Beek, and Walsh]{handbook}
Francesca Rossi, Peter van Beek, and Toby Walsh.
\newblock \emph{Handbook of Constraint Programming {(Foundations} of Artificial
  Intelligence)}.
\newblock Elsevier Science Inc., 2006.
\newblock ISBN 0444527265.

\bibitem[Sabin and Freuder(1996)]{sabin_configuration_1996}
Daniel Sabin and Eugene~C. Freuder.
\newblock Configuration as composite constraint satisfaction.
\newblock \emph{Proceedings of the (1st) Artificial Intelligence and
  Manufacturing Research Planning Workshop}, pages 153--161, 1996.

\bibitem[Smith(1990)]{kids}
Douglas~R. Smith.
\newblock {KIDS} - a {Knowledge-Based} software development system.
\newblock In \emph{Automating Software Design}, pages 483--514. {MIT} Press,
  1990.

\bibitem[Stumptner et~al.(1998)Stumptner, Friedrich, and
  Haselb{\"o}ck]{stumptner_generative_1998}
Markus Stumptner, Gerhard~E. Friedrich, and Alois Haselb{\"o}ck.
\newblock Generative constraint-based configuration of large technical systems.
\newblock \emph{Artif. Intell. Eng. Des. Anal. Manuf.}, 12:\penalty0 307--320,
  1998.

\bibitem[Xu et~al.(2008)Xu, Hutter, Hoos, and {Leyton-Brown}]{satzilla}
L.~Xu, F.~Hutter, H.~H. Hoos, and K.~{Leyton-Brown}.
\newblock {SATzilla:} portfolio-based algorithm selection for {SAT}.
\newblock \emph{J. Artif. Intell. Res. {(JAIR)}}, 32:\penalty0 565--606, 2008.

\end{thebibliography}

\end{document}